\documentclass[5p,times]{elsarticle}

\usepackage{lineno,hyperref}
\modulolinenumbers[5]

\journal{Journal of Robotics and Autonomous Systems}

\usepackage{amsmath}
\usepackage{colortbl}
\usepackage{multirow}
\usepackage{colortbl}
\usepackage[nolist]{acronym}
\usepackage[font=footnotesize]{subfig}
\usepackage{commath}

\def\eg{\emph{e.g.}} 
\def\ie{\emph{i.e.}}

\def\etal{\emph{et al.}}

\begin{acronym}
\acro{CNN}{Convolution Neural Network}
\acro{FCN}{Fully Convolutional Network}
\acro{RPN}{Region Proposal Network}
\end{acronym}

\begin{document}

\begin{frontmatter}

\title{Where is my hand? Deep hand segmentation for visual self-recognition in humanoid robots}

\author{Alexandre Almeida\fnref{ist}}
\author{Pedro Vicente\fnref{ist}\corref{mycorrespondingauthor}}
\cortext[mycorrespondingauthor]{Corresponding author}
\ead{pvicente@isr.tecnico.ulisboa.pt}
\author{Alexandre Bernardino\fnref{ist}}
\fntext[ist]{A. Almeida, P. Vicente,  A. Bernardino \\
            Institute for Systems and Robotics (ISR/IST), LARSyS, \\
            Instituto Superior Tecnico, University Lisboa, Avenida  \\
            Rovisco Pais 1 - Torre Norte, 1049-001 \\
            Lisboa, Portugal \\}

\begin{abstract}
The ability to distinguish between the self and the background is of paramount importance for robotic tasks. The particular case of hands, as the end effectors of a robotic system that more often enter into contact with other elements of the environment, must be perceived and tracked with precision to execute the intended tasks with dexterity and without colliding with obstacles. They are fundamental for several applications, from Human-Robot Interaction tasks to object manipulation. Modern humanoid robots are characterized by high number of degrees of freedom which makes their forward kinematics models very sensitive to uncertainty. Thus, resorting to vision sensing can be the only solution to endow these robots with a good perception of the self, being able to localize their body parts with precision.
In this paper, we propose the use of a \ac{CNN} to segment the robot hand from an image in an egocentric view. It is known that \ac{CNN}s require a huge amount of data to be trained. To overcome the challenge of labeling real-world images, we propose the use of simulated datasets exploiting domain randomization techniques.
We fine-tuned the Mask-RCNN network for the specific task of segmenting the hand of the humanoid robot Vizzy.
We focus our attention on developing a methodology that requires low amounts of data to achieve reasonable performance while giving detailed insight on how to properly generate variability in the training dataset. Moreover, we analyze the fine-tuning process within the complex model of Mask-RCNN, understanding which weights should be transferred to the new task of segmenting robot hands. 
Our final model was trained solely on synthetic images and achieves an average IoU of $82\%$ on synthetic validation data and $56.3\%$ on real test data. 
These results were achieved with only 1000 training images and 3 hours of training time using a single GPU.

\end{abstract}


\end{frontmatter}


\section{Introduction}
\label{sec:intro}

\begin{figure}[h!]
    \centering
    \begin{tabular}{@{}c@{}}
    \subfloat    []{
    \includegraphics[width=0.49\linewidth]{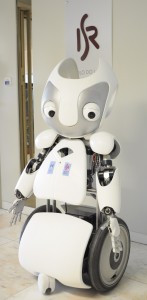}
    } 
    \end{tabular}
    \begin{tabular}{@{}c@{}}
    \subfloat    []{
    \includegraphics[width=0.45\linewidth]{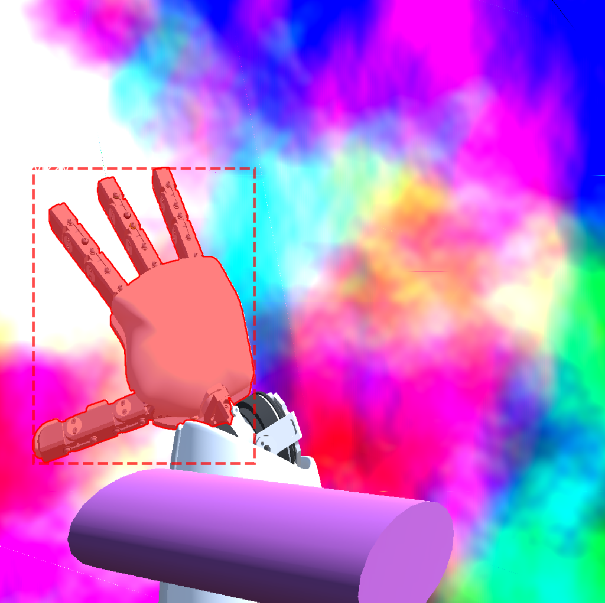}
    } \\
    \subfloat    []{
    \includegraphics[width=0.45\linewidth]{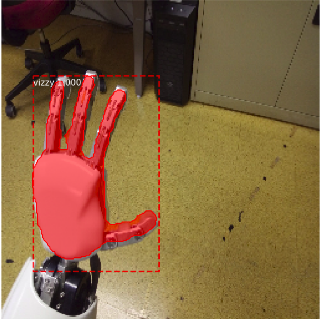}
    }
    \end{tabular}
    \caption{Some examples of hand segmentation. (a) our Vizzy Humanoid Robot. (b) a sample from the dataset of synthetically generated robot views in a graphics simulator (this dataset is used in training and the ground truth masks are generated automatically) and (c) a sample from the real dataset, where images are captured by the robot cameras. This set is used for testing. The observed mask overlaid in the image was produced by our algorithm. }
    \label{fig:cover}
\end{figure}

The ability to recognize themselves and distinguish between self and background using visual sensing is an important skill for intelligent agents. Indeed, the perception of the self, \textit{i.e.}, the ability of detecting and identifying its own body (limbs position and physical boundaries), is beneficial not only for self-centered-actions (\textit{e.g.},  pick, place or pushing)  but also for interaction with other agents (\textit{e.g.}, hand-shaking or collaborative tasks). 
In humanoid agents (humans and humanoid robots), the hands are an important part of the body for tasks like grasping, reaching and environment exploration, hence having a precise localization of the own hands with respect to the other items of the environment is of paramount importance. Moreover, hand self-recognition \textit{per se} can be also exploited as a pre-processing step for features extraction (\textit{e.g.}, for learning hand affordances, in \cite{saponaro2017icdl}) or to use it on calibration procedures (\textit{e.g.}, as proposed in \cite{vicente2016jint}). 
Visual self-recognition of the robot hands can be perform using several strategies.
For instance, 2D hand keypoints can be used as self-recognition. However, 
they only provide information of part of the hand (usually the phalanx locations). While that can be
enough for some tasks, \textit{e.g.}, hand gestures recognition, it might not be enough for other tasks that require full knowledge about the spatial layout of the hand, \textit{e.g.}, object grasping.
For this reason, we want to explore hand segmentation as a way of visual self recognition. Figure~\ref{fig:cover} illustrated the purpose of our work: to segment all the pixels belonging from a real robot hand (in our case the Vizzy Humanoid robot), using single RGB images extracted from the robot cameras. We adopt a supervised learning approach using training data acquired solely from graphical simulations.

In the literature, some works exploit motions generated by issuing suitable motor commands to perform hand segmentation in robots (\cite{laflaquiere2019icdl} \cite{lanillos2017tcds}). In \cite{laflaquiere2019icdl}, the authors propose to learn the body image of the robot Pepper by segmenting the robot body from the background. A neural network is trained receiving a motor input (four joints of the arm) and outputs the estimated body image and body segmentation. However, it is assumed that the kinematics and appearance models of the robot are accurate and do not change over time. 
In another work \cite{lanillos2017tcds}, it is employed a self-perception scheme, where the combination of motor commands and visual perception is used to perform self-detection. Lanillos \textit{et al.} \cite{lanillos2017tcds} propose to explore motor contingencies to segment the robot hand.
However, the robot should move to perform a good body segmentation/detection.

In a real-world scenario, humanoid robots are, usually, not perfectly calibrated since their structure suffers from wear and tear, flexibility of materials, and accumulation of errors due to long kinematic chains. Therefore, it is crucial to exploit the robot's perception and avoid relying entirely on open-loop motor commands to perform tasks. 
Indeed, vision or tactile perception was used in \cite{vicente2016Frontiers,zenha2018icdl,stepanova2019ral}, to correct the kinematics models and improve precision in task execution. Other approaches propose to guide the robotic arm through visual feedback control (\cite{vicente2017icra,fantacci2017iros,nascimento2020icarsc}). 
In both cases, self-perception of the robot end-effectors (\textit{e.g.}, using vision or tactile sensing) is essential to provide good accuracy in the given task.

The problem of segmenting humanoid robot hands can gain inspiration in the problem of segmenting a human hand. The main differences between these two problems are: (i) many human hand segmentation methods rely on skin-color information \cite{Kakumanu2007, Jones2002}, which perform poorly on cluttered domains with different lighting effects, and (ii) there is a wide variety of human hand datasets available and already annotated \cite{Khan2018} whilst, to our knowledge, there is no pre-existent annotated dataset of humanoid robotic hands in egocentric view.

As far as we know, there are not many works performing robotic hand segmentation exploiting vision only. A notable exception is \cite{Leitner2013}, where J. Leitner \textit{et al.} propose a genetic algorithm to detect the hands of two different humanoid robotic models. They also propose two approaches for detection: (i) detect the fingertips of the robots' hands, and (ii) detect the full hand. The second task is more desirable since it provides more information, but it is described as a more complicated task, requiring ten times more training time than the first approach. However promising, that work does not present numerical results due to the lack of suitable datasets, making it hard to compare it with our work.  

CNNs (Convolutional Neural Networks) recently started to solve many problems in computer vision, due to their high efficiency in extracting features from images. These features can be used to classify objects and make bounding box predictions for each object. A recent work in this area, Mask RCNN \cite{R-cnn2017}, achieved great accuracy at these two tasks, while also retrieving a binary mask pixel-wise segmentation for each object. The advantage of using this type of network is that no feature engineering is required (feature extraction is part of the deep learning process). Recent improvements in parallel programming with GPUs (\textit{e.g.}, CUDA\footnote{CUDA webpage \url{https://developer.nvidia.com/cuda-zone}}) make deep learning algorithms a good choice for many computer vision problems. Moreover, the weights trained on large datasets (\textit{e.g.}, COCO \cite{Lin2014}, ImageNet \cite{Russakovsky2014}, etc) are often made available to the research community and can be used to initialize the weights of networks to be trained for different tasks. This often leads to improvements in training times and generalization when compared to methods that initialize weights randomly. 

Our goal with is to propose a deep learning method to segment the hand of a robot from the background. To achieve this task, we have to overcome two major challenges, namely: (i) how to collect large amounts of training data, given the absence of any pre-existent datasets, and (ii) how to fit a model, pre-trained on a general purpose dataset, to our specific domain (indoor cluttered domain, with robotic hands). In order to address (i), we propose to use and explore domain randomization \cite{Tobin2017}, a method that relies on training a model with exclusively simulated data aiming to retain most of the performance, in real data. With this method we do not need to collect and annotate images manually. To address (ii), we propose to fine-tune the pre-trained weights\footnote{the parameters that are adjusted during training through backpropagation, to decrease the training loss (\textit{e.g.} the kernels/filters weights)} and the \textit{hyperparameters}\footnote{any parameter related to the training process, that can be tuned manually without having to make any change in the network's architecture and are not changed during a training epoch (\textit{e.g.} learning rate, loss functions, number of train/val epochs)} of a State-of-the-art \ac{CNN}, in particular, the Mask RCNN network.

The rest of the article is organized as follows. In Section~\ref{sec:backg} we will revise the background on deep learning approaches for image segmentation. In Section~\ref{sec:Methodology} we will present our methodology to solve the problem of visual self recognition via hand segmentation. Then, in Section~\ref{sec:exp_setup} the implementation details, the datasets and the evaluation metrics are presented. In Section~\ref{sec:results}, we present our experimental results on simulation and on the real robot - Vizzy. Finally, in Section~\ref{sec:concl}, we draw our conclusions and sketch future work directions.

\section{Overview of Image Segmentation using Deep Learning}
\label{sec:backg}

\ac{FCN} \cite{Long2014} is a state-of-the-art \ac{CNN} for pixel-wise classification tasks (\ie, segmentation), without the need for any pre or post-processing (\eg, superpixels, bounding box proposals, like in \cite{Farabet2013, Hariharan2014}). This method extends previous classification networks to the task of semantic segmentation by up-sampling final layers to match the input image size, using deconvolutional layers \cite{Dumoulin2016}. These layers do not need to be fixed to perform bilinear interpolation, instead, they can learn how to up-sample optimally.  

U-Net \cite{LinMS2016} builds on the \ac{FCN} \cite{Long2014}. It takes an image as input, down-samples it to lower dimensions through convolution and max pooling operations in order to convert the original image into feature vectors. This step is followed by up-sampling paths that apply transpose convolutions \cite{Dumoulin2016} to recover and map the results back to the dimensions of the original image. However, differently from \ac{FCN} \cite{Long2014}, the authors propose a modification in the up-sampling part. They propagate context information to higher resolution layers through concatenating feature maps from the corresponding contracting path. This type of network was designed for biomedical applications (\eg segmentation of cells), making it an ideal choice to segment images with low background clutter and clear boundaries between different objects (no overlapping).



RefineNet \cite{RonnebergerFB15} is a multi-path network, \ie, it exploits features at multiple resolutions to achieve high-resolution segmentation. Also, this network uses ResNet \cite{He2015} models, pre-trained on ImageNet \cite{Russakovsky2014}, for feature extraction.


Other deep learning approaches to the segmentation problem rely on making candidate object proposals and then classify each proposal separately \cite{Pinheiro2016, Pinheiro2015, R-cnn2017} (instead of classifying the whole image at once). In the methods proposed in \cite{Pinheiro2016, Pinheiro2015}, segmentation precedes recognition, which is slower and less accurate, according to more recent studies \cite{R-cnn2017}. Instead, He \etal\ propose Mask R-CNN  which is divided into two stages. 
%
%
The first stage is called the \ac{RPN} \cite{Girshick2015} to generate candidate object proposals. It generates anchors of different sizes and ratios, centered in every pixel of the feature map and classify each anchor as background or foreground. Each anchor classified as foreground is accepted for the second stage, which consists in assigning each anchor to its final class. A bounding box regression step predicts an adjustment to the anchor (for the bounding box prediction) and retrieves a binary mask of the object present in the anchor. These three tasks (\textit{i.e.}, bounding box regression, classification and segmentation) are done in parallel, in order to increase speed. 

To our knowledge, Mask R-CNN \cite{R-cnn2017} is, overall, the most successful state-of-the-art deep learning architecture for tasks of both object recognition and segmentation, showing clear improvements over previous methods, both in terms of speed and accuracy. Moreover, Mask R-CNN models pre-trained on COCO dataset \cite{Lin2014} are available online \footnote{Mask R-CNN pre-trained model's weights, on COCO dataset: \url{https://github.com/matterport/Mask\_RCNN/releases/download/v2.0/mask\_rcnn\_coco.h5}} making this the architecture upon which we build our work.
\section{Methodology}
\label{sec:Methodology}

We propose to extract the bounding box and a binary mask of the instances of humanoid robotic hands, present in an input image. For the neural network architecture, we use Mask R-CNN \cite{R-cnn2017} with the ResNet-101-FPN \cite{He2015, Lin2016} backbone, for the feature extraction process. 

\begin{figure}
    \centering
    \includegraphics[width=1\linewidth]{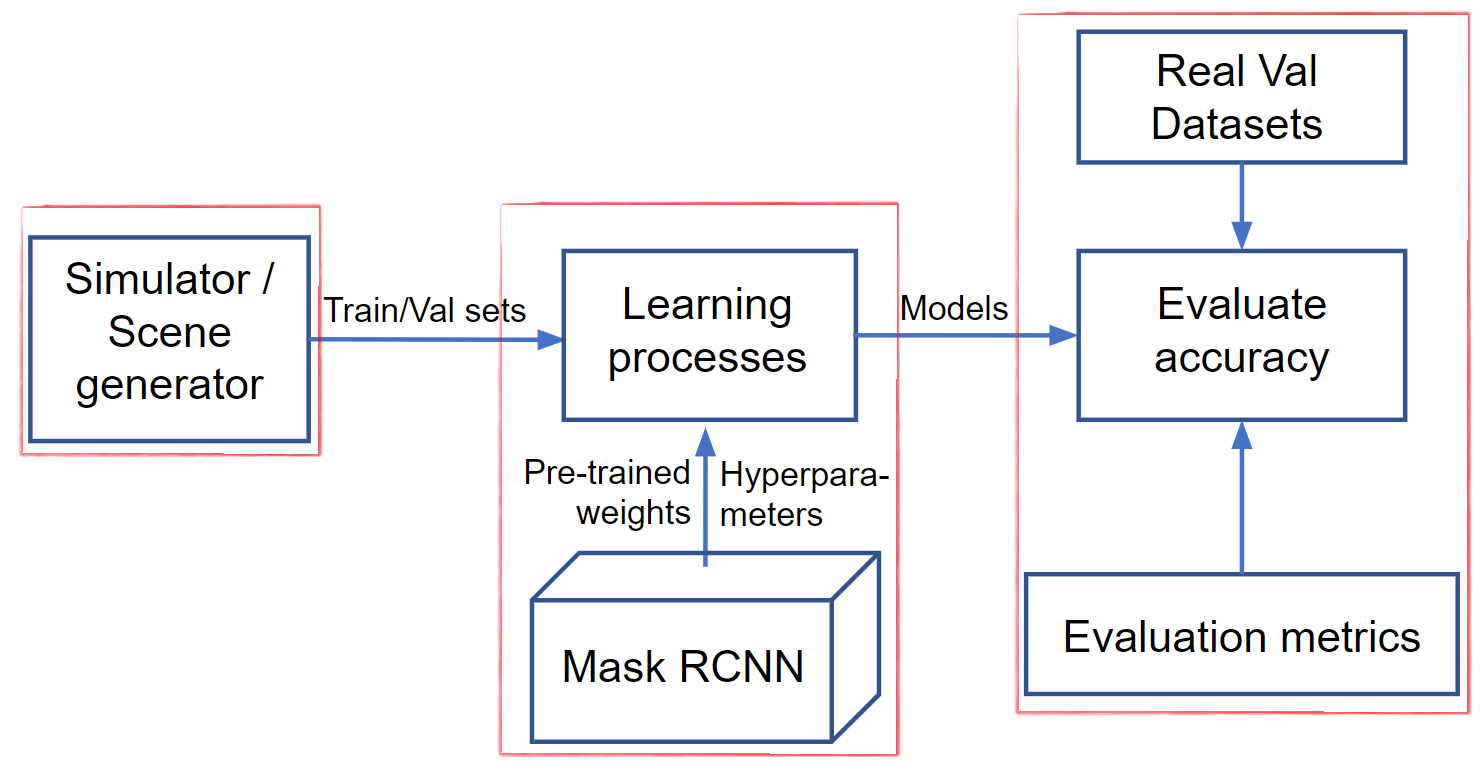}
    \caption{Overview of the methodology pipeline, using the architecture of Mask R-CNN \cite{R-cnn2017}. Our methodology is divided in three main blocks: i) scene generator, ii) learning process and iii) evaluation}
    \label{fig:method_pipeline}
\end{figure}

Our methodology can be divided into three main blocks (see Figure~\ref{fig:method_pipeline}). The scene generator block generates the training and validation (simulated) data. We generate an RGB image and the corresponding ground-truth bounding box and mask. This data is used to feed the network during training. The learning process block encapsulates the strategies to adapt to a different classification task a \textit{model} that was pre-trained on a dataset of a significantly different domain. The last block is the evaluation, where we evaluate the loss and accuracy of trained \textit{models}, using appropriate metrics and real validation datasets that will be defined in Section \ref{sec:exp_setup}.

\subsection{Mask R-CNN architecture}
\label{sec:mask_r-cnn_architecture}

\begin{figure}
    \centering
    \includegraphics[width=1\linewidth]{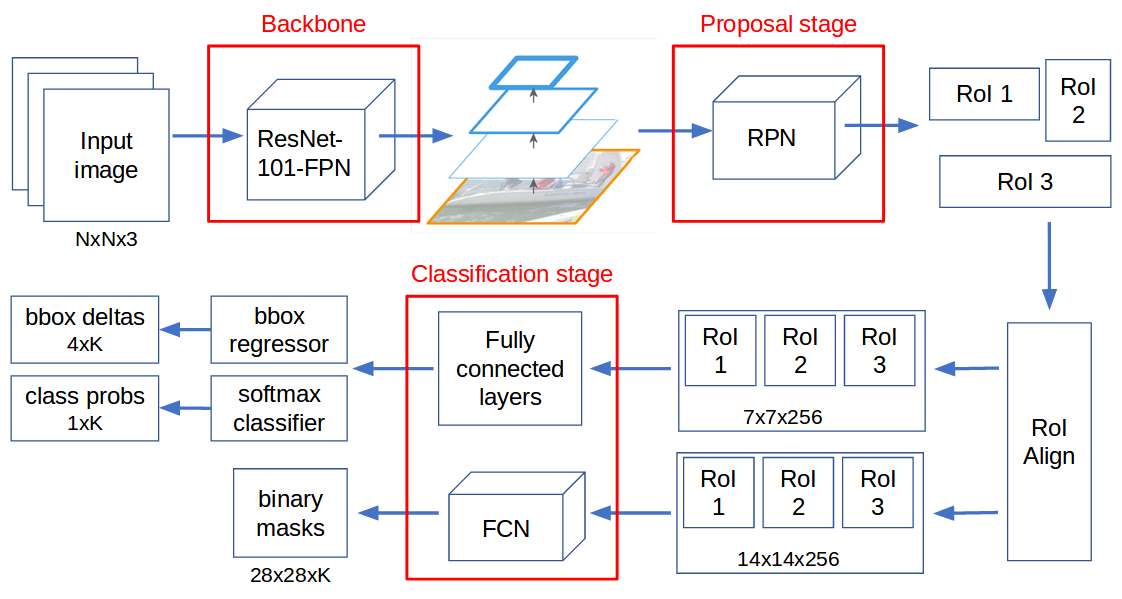}
    \caption{Overview of the architecture of Mask R-CNN \cite{R-cnn2017}, divided by its 3 main blocks.}
    \label{fig:red_blocks}
\end{figure}

In Figure \ref{fig:red_blocks}, we provide an overview of the architecture that was used for this work. We divided it into three main stages and will give a brief explanation on each of them.

The first stage is called the backbone, where we feed the input images into a residual neural network (the ResNet-101-FPN \cite{He2015, Lin2016}). This stage has the task of learning to extract the most relevant features from the input image and aggregate this information into feature maps, that will be used for classification. Furthermore, the backbone is divided into four blocks of layers (called levels), where the last layer of each block is the feed-forward connection to a different level of the Feature Pyramid Network (FPN). More in detail, a $1 \times 1$ convolutional layer is applied to the output of each block of the backbone and the result is added to the feature map of the upper layer (except for the layer of the last block).  Finally, a $3 \times 3$ convolutional layer is applied to each level, resulting in four different feature map levels.

The proposal stage (the second stage) has the task of learning to propose regions of interest from different levels of the FPN. For each pixel of the feature maps, a total of 15 anchors, of different scales and ratios, are centered in that pixel. Each anchor is classified as background or foreground and a bounding box regression is made to adjust the positive anchors to their respective groundtruth bounding boxes. From all the positive regions of interest, the network chooses $N$ regions to be passed into the next stage, where $N$ is a \textit{hyperparameter}.  Out of these $N$ regions of interest, there is a fixed ratio of negatives (\textit{i.e.}, regions of interest with a score lower than 0.3) and positives (\textit{i.e.}, regions of interest with a score greater than 0.7), also set as \textit{hyperparameters}. This means that all images implicitly have negative samples, so there is no need for purely negative images, during training. This stage has 2 loss functions, (i) $rpn\_class\_loss$ accounts for the error in predicting the class labels of each anchor, and (ii) $rpn\_bbox\_loss$ accounts for the error in predicting the deltas\footnote{deltas are the bounding box regression parameters. Two for the displacement of the bounding box center (x and y) and two for the error in height and width} 
that best adjust the anchor to its corresponding groundtruth bounding box. For the $rpn\_class\_loss$, the cross-entropy loss is used (Equation~\ref{eq:cross_ent_loss}), where $K$ represents the total number of classes (in our case $K=2$), $y_i \in \{0, 1\}$ the true value of class $i$ and $\hat{y_i} \in \{0, 1\}$ the predicted value of class $i$. For the $rpn\_bbox\_loss$ the Huber loss is used (Equation~\ref{eq:huber_loss}), where $z \in \mathbb{R}^4$ represents the true deltas' vector and $\hat{z} \in \mathbb{R}^4$ the predicted deltas' vector.

\begin{align}
    \text{CE}(y, \hat{y}) &= - \sum_{i=1}^{K} y_i \log(\hat{y_i}) \label{eq:cross_ent_loss} \\
    \text{Huber}(z, \hat{z}) &= 
    \begin{cases} 
    \frac{1}{2} \cdot \norm{z - \hat{z} }^2, & \mbox{if }  \abs{ z - \hat{z} } < 1 
    \\ \abs{ z - \hat{z} } - \frac{1}{2}, & \mbox{if } \text{otherwise} \end{cases} 
    \label{eq:huber_loss}
\end{align}  

Finally, for the classification stage, all the proposed regions of interest are picked and down-sampled to a fixed dimension ($7 \times 7$ for the classification and bounding box regression branches and $14 \times 14$ for the mask branch). This stage predicts 3 outputs, (i) the final bounding box deltas, with dimension $4 \times K$, (ii) the final class labels $1 \times K$ vector, and (iii) a $28 \times 28 \times K$ binary mask.
Each output has a loss function associated; output (i) uses the Huber loss (Equation \ref{eq:huber_loss}), output (ii) uses the cross-entropy function (Equation \ref{eq:cross_ent_loss}), and output (iii) also uses the cross-entropy to compute a per-pixel loss. Since we are changing the classification task to a binary classification, the output layers of this stage were changed and cannot be initialized with the weights of the pre-trained \textit{model}. Therefore, these layers must be trained from scratch.

\subsection{Data generation}
\label{sec:Data_gen}

To our knowledge, at the beginning of this work there was no dataset of humanoid robotic hands with annotated ground-truth masks, available for research. Given the considerable amount of time required to collect and annotate enough training and validation images, we propose to build our own custom dataset in simulation, using a 3D model of a humanoid robotic hand (see Figure~\ref{fig:cover} for an example of the simulated and real hand). We apply domain randomization \cite{Tobin2017} to generate a dataset with enough variability to overcome the \textit{reality gap}, \textit{i.e.}, to achieve reasonable performance when validating in a real scenario a \textit{model} that was trained only with simulated images. 

In the simulation domain, we randomize properties of the scene related to the background, foreground and lighting effects. In this work, we will focus largely on randomizing properties related to the two former components, since they show greater improvements for the overall accuracy, in the real domain.

To provide variability in the background, we consider two components, distractor objects, and background wall. As distractor objects, we consider four types of 3D parametric shapes (ellipsoids, parallelepipeds, elliptic cylinders, and spherocylinders) with randomized shapes, sizes, and colors (solid colors only). For the background wall, we generate three types of textures: solid random colors, Perlin noise \cite{Ken2002}, and real backgrounds. 

\begin{figure}
    \centering
    \includegraphics[width=.7\linewidth]{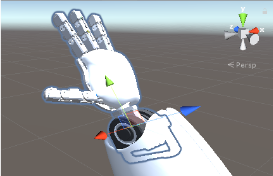}
    \caption{Reference frame of the wrist joint. The red, green and blue axes correspond to the $x, y, z$ axes, respectively.}
    \label{fig:hand_orientation}
\end{figure}

In the foreground, we randomize the properties of the 3D hand model. This includes the position and orientation of the arm (which contains the hand and wrist), and the orientation of the hand relative to the arm. In Figure \ref{fig:hand_orientation}, we show the reference frame of the wrist joint. The initial configuration for the orientation of the hand, relative to the arm, is $\gamma_{x,y,z} = [0, 0, 0]$, in angular units, which means the hand starts aligned with the reference frame of the wrist.  

\begin{figure*}
    \centering
    \includegraphics[width=0.8\linewidth]{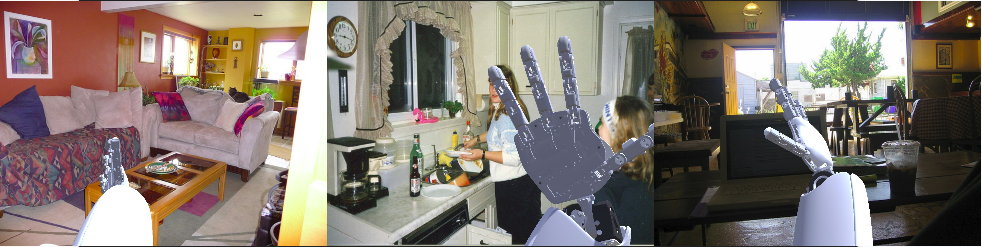}
    \caption{3 examples of images using real backgrounds, from the COCO dataset.}
    \label{fig:negative_images}
\end{figure*}

In the context of this work, the real robot has physical limitations that restrict the relative orientation of its own hand with respect to its arm. If we consider the orientation of the hand 
as 3 separated randomization parameters (one parameter for each axis), the possible configurations for these parameters are bounded by the limits of the real robot. In other words,  

\noindent$[\gamma_1^{\text{lower}}, \gamma_2^{\text{lower}}, \gamma_3^{\text{lower}}]$ and $[\gamma_1^{\text{upper}}, \gamma_2^{\text{upper}}, \gamma_3^{\text{upper}}]$ are the lower and upper bounds for the angles of these 3 degrees of freedom. 
For this reason, we can define ranges for the values of each orientation parameter that avoid unfeasible configurations in the target domain. Indeed, configurations unfeasible in the real domain would not bring relevant data to the learning process. 
Thus, we can increase variability on the hand orientation without decreasing the performance of the model, since we have access to prior knowledge about the domain.

As for the lighting effects, we generate 3 spotlights and randomize the position and orientation of each one. Note that, in our implementation, we do not randomize the camera properties but, since the position and orientation of all objects are randomized, we capture different perspectives of the scene. 

The target domain (real domain) of our work may contain high clutter, \textit{i.e.}, a high amount of different colors, overlapping objects, and different lighting effects, around the objects of interests (the hands). Even using the domain randomization technique, some aspects of the real-world are difficult to generate and to overcome using only simulation environments. 
This condition can generate a lower accuracy when evaluating the model. To help address this problem, we also use real backgrounds in our training and validation simulated datasets. To do this, we use a number of images belonging to the super-category \textit{indoor} of the COCO \cite{Lin2014} dataset and overlap on top of each image a randomized instance of the 3D robot hand model. In Figure~\ref{fig:negative_images}, we provide 3 examples of images with real backgrounds used for the training and validation simulated sets.

To generate the ground-truth synthetic data, we automatically extract a binary mask of each synthetic image (where $1$ corresponds to hand pixels and $0$ to background pixels), from the simulator. The bounding box coordinates can be extracted from the mask. Regarding real data for the evaluation, we collected images from different real scenarios (described further, in Section \ref{sec:test_sets}) and extracted the ground-truth data from each real image manually, using an annotation tool\footnote{Custom annotation tool GitHub \url{https://github.com/alexalm4190/Mask_RCNN-Vizzy_Hand/blob/master/utils/annotate_masks.py}} developed from scratch.

\subsection{Learning}
\label{sec:learning}

During training, the \textit{weights} are adjusted using only data generated in simulation. 
According to J. Yosinski \textit{et al.} \cite{YosinskiCBL14}, it is possible to retain the generalization ability of \textit{models} trained in large scale datasets, even after fine-tuning the \textit{weights} to perform different classification tasks, in different domains. This can be achieved by adopting a good training strategy, \textit{i.e.} choosing the correct layers to freeze and to fine-tune, during the learning process. 

It is commonly accepted that initial layers extract features that represent more general aspects of the dataset thus, the \textit{weights} tend to not change much in these layers, even when the dataset or the classification task are  changed. Therefore, initializing these \textit{weights} from already learned values, will improve the convergence.
Following this principle, we initialize the network with \textit{weights} pre-trained in the COCO \cite{Lin2014} dataset, except for the output layers that predict the classes, bounding boxes and masks, because we change the number of output classes to 2. The training and validation datasets used are generated in simulation, following the process explained in Section \ref{sec:Data_gen}. We sample, approximately, a total of $1000$ images from the simulator, in which $80\%$ are used for training and the remaining $20\%$ are used for validation. 

In the Mask R-CNN \cite{R-cnn2017} architecture, the total loss is divided into five losses. Two are to account for the loss when proposing a region of interest that contains an object, in the RPN \cite{Ren2015} stage. The \textit{rpn\_class\_loss} is for the predicted binary class, object/not object, and the \textit{rpn\_bbox\_loss} is for the predicted bounding box that contains the object. The three remaining losses account for the predictions of the final class of the object (\textit{mrcnn\_class\_loss}), hand or background, the final predicted offsets for the bounding box (\textit{mrcnn\_bbox\_loss}) and the predicted binary mask (\textit{mrcnn\_mask\_loss}). 
The training loss is defined as the sum of the five losses (with equal contribution to the total loss) and, for each learning process, we also monitor the sum of the five losses on the validation set. We apply early stopping, to avoid overfitting in the training process, by monitoring the total validation loss, with a threshold of 15 epochs. Then, the final model is chosen based on the lowest validation loss, during that training episode. We also set a maximum of 150 training epochs, in case the loss does not stop decreasing.
\section{Experimental setup}
\label{sec:exp_setup}

In this section, we will give some details on the platforms used to implement our processes. We also present the datasets produced to train, validate and test our method as well as the evaluation metrics.

\subsection{Implementation details}
\label{sec:imp_details}

To implement the data generation process, we use Unity\footnote{Unity webpage \url{https://unity.com/}}, a game development platform. 
Unity has a higher graphics quality, compared with the usual robotic simulators (\eg, Gazebo), since it was devised to develop games.
In our case, we do not care about physics or good simulation in terms of robot motion control. Thus, Unity is suitable for this project since we will be able to produce better quality images. Moreover, it also allows to generate images at a higher rate, compared with other 3D simulation platforms. 
To render the robotic hand in the images, we import a 3D model of Vizzy \cite{Plinio2015}, a humanoid robot designed for assistive robotics. Some of the physical restrictions of the robot are also implemented in Unity, like the ranges of angular movements in some joints (forearm pronation, wrist abduction, and wrist flexion, as specified in \cite{Plinio2015}). This is done to avoid unfeasible hand poses that are not coherent with the ranges of the real robot.

We noticed that only using the hand (without a visible arm attached) leads to poor results in segmentation in the real domain (check Section~\ref{sec:dr_results} for more details). 
So, to improve the results, we also attached the 3D model of the arm to the hand, in Unity. Since the rest of the body is usually not visible in the real domain, we do not use it in simulation.

For the learning process, we exploit an open-source implementation of Mask R-CNN\footnote{Mask R-CNN GitHub \url{https://github.com/matterport/Mask_RCNN}}, that follows the architecture of the original work \cite{R-cnn2017}.

We have a public GitHub repository\footnote{Public GitHub repository, with our project \url{https://github.com/alexalm4190/Mask_RCNN-Vizzy_Hand}} with the changes made to the original Mask R-CNN repository and all the code used and created to implement our methods and experiments.

\subsection{Datasets}
\label{sec:test_sets}
In this section, we will present the datasets used to train, validate, and test our method. The data was divided into two parts: the Simulated Dataset composed of synthetic images generated using Unity, and Real-World Dataset composed of images acquired with the real robot.
We exploit the simulated datasets for training and validation of our method and the real-world datasets for validation and test. 

\subsubsection{Simulated Datasets}
\label{sec:simdatasets}

We consider 6 dataset partitions that incrementally add the features under analysis,
\ie, each dataset includes images from the previous described dataset. 
Set A (SolidBg hand) is the baseline, having solid colors and only the hand in the foreground. Set B (SolidBg arm) adds the arm attached to the hand. Set C (Perlin noise) adds Perlin noise to create backgrounds with texture diversity. Set D (Distractor objects) adds distractor objects to the foreground. Set E (Lighting effects) adds lighting effects. Finally, Set F (Real backgrounds) replaces some of the synthetic backgrounds with real ones.
Each simulated datasets has, approximately, 1000 images. 
By training the network with the different sets we can analyze the influence of the different features in the data generation process. In particular, we are interested in understanding the influence of: i) the attached arm, ii) randomization of the background texture and color - by using solid, Perlin noise or real backgrounds, iii) partial occlusion of the hand by using distractor objects and iv) light in the image generation process.

\begin{figure}
    \centering
    \includegraphics[width=\linewidth]{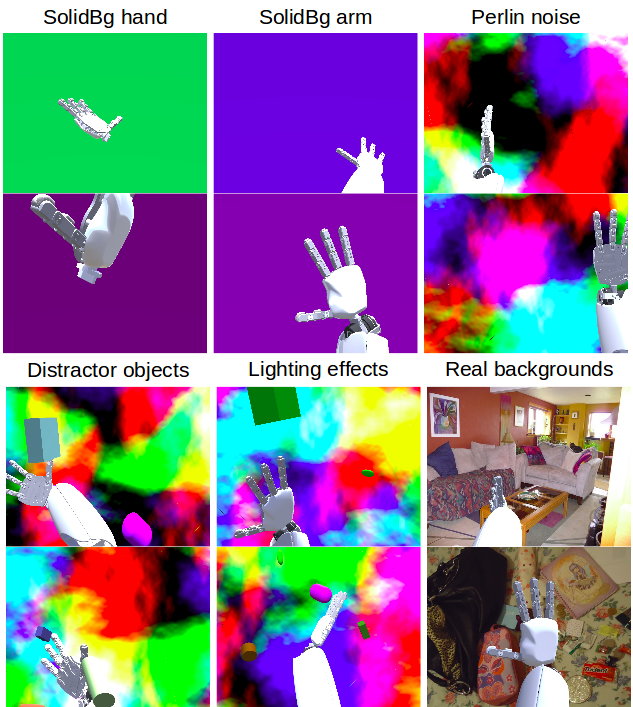}
    \caption{Datasets generated in Unity, using domain randomization.}
    \label{fig:dr_datasets}
\end{figure}

In Figure \ref{fig:dr_datasets}, we show images from each of the six datasets, generated in Unity. Note that each different feature (except for the arm attachment) is added to the others and does not replace previous features, \textit{e.g.} in the \textit{Perlin noise} dataset we have images with Perlin noise backgrounds and other images with solid colored backgrounds (each taking up, approximately, half the dataset). 

The simulated datasets are divided into 80\% training and 20\% validation.

\subsubsection{Real-World Datasets}

\begin{figure*}
    \centering
    \subfloat    []
    {\includegraphics[width=0.6\linewidth]{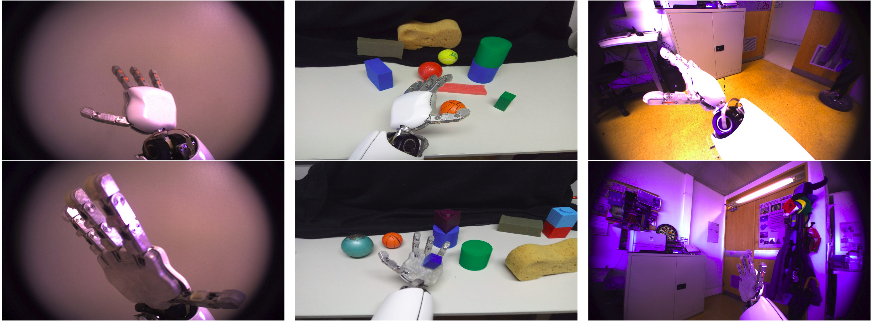}}
    \subfloat   [] {
    \includegraphics[width=0.15\linewidth]{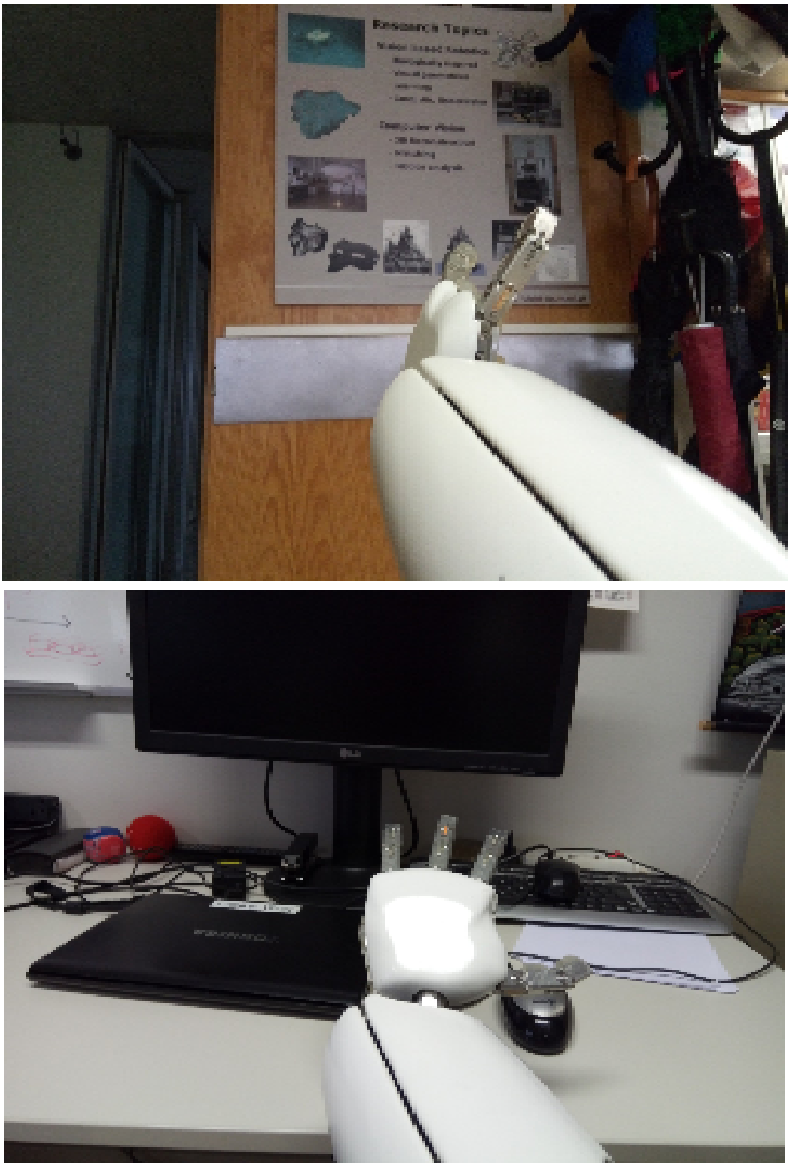}}
    \caption{Real Datasets. Validation Sets in (a): Low cluttered (left), medium cluttered (center) and high cluttered (right) backgrounds. And the test set in (b).}
    \label{fig:test_sets}
\end{figure*}

To validate and test the accuracy of the models in the real domain, we collected and annotated images of the real robot's hands in four different scenarios: (i) low clutter background, (ii) medium clutter background, (iii) high clutter background and (iv) test set scenario.

The low cluttered background dataset (i) is composed of a simple solid background with different hand positions and orientations, with the purpose of validating the ability of the model to detect the hand, without much interference from background textures. This allows us to evaluate how similar
the real robot’s hand is to the 3D model, that is used in Unity to generate the positive instances of the training data. If the model learns to detect almost as accurately in this dataset as in the simulated dataset, it means the 3D model has the appropriate textures, colors, size and shape to represent the hand of the real robot, in simulation. 

For the medium cluttered background dataset (ii) we setup a scene with a relatively simple background, but with multiple distractor
objects that, sometimes, occlude part of the hand. We vary the configuration of the objects, in a tabletop scenario and the pose of the robot’s hand and arm. 

For the high cluttered dataset (iii) we take the images in an office-like environment, with high amounts of clutter in the background and different lighting
effects. We vary the orientation of the cameras (to capture different parts of the environment) and the pose of the hand and arm of the robot. This dataset allows us to evaluate how the model performs in extreme situations, with real textures very different from the textures generated in the simulator. 

Finally, the test set scenario (iv) was collected in a scenario with the same level of difficulty of (iii) - the high clutter validation dataset. With this dataset, we are able to test our final model’s capacity to generalize to data that was neither used to fine-tune the network's \textit{weights} and \textit{biases}, during training, nor the model's \textit{hyperparameters}, during validation. 

For a better understanding of the Real-World environments, we provide two images of each dataset. Examples of the validation sets can be seen in Figure~\ref{fig:test_sets} (a). In the left column is the low cluttered dataset (i). In the center of Figure~\ref{fig:test_sets} (a), is the medium cluttered dataset (ii) and in right side of Figure~\ref{fig:test_sets} (a) is the higher clutter dataset (iii). Finally, examples of the test set (iv) are given in Figure~\ref{fig:test_sets} (b).

It is important to emphasize, that the Real-World datasets are smaller than the simulated ones (around 20 to 30 images each), since it is more difficult to generate the hand segmentation in the real images (it is usually a manual and long process). 

\subsection{Evaluation metrics}
\label{sec:eval_metrics}

\begin{figure}
    \centering
    \includegraphics[width=0.8\linewidth]{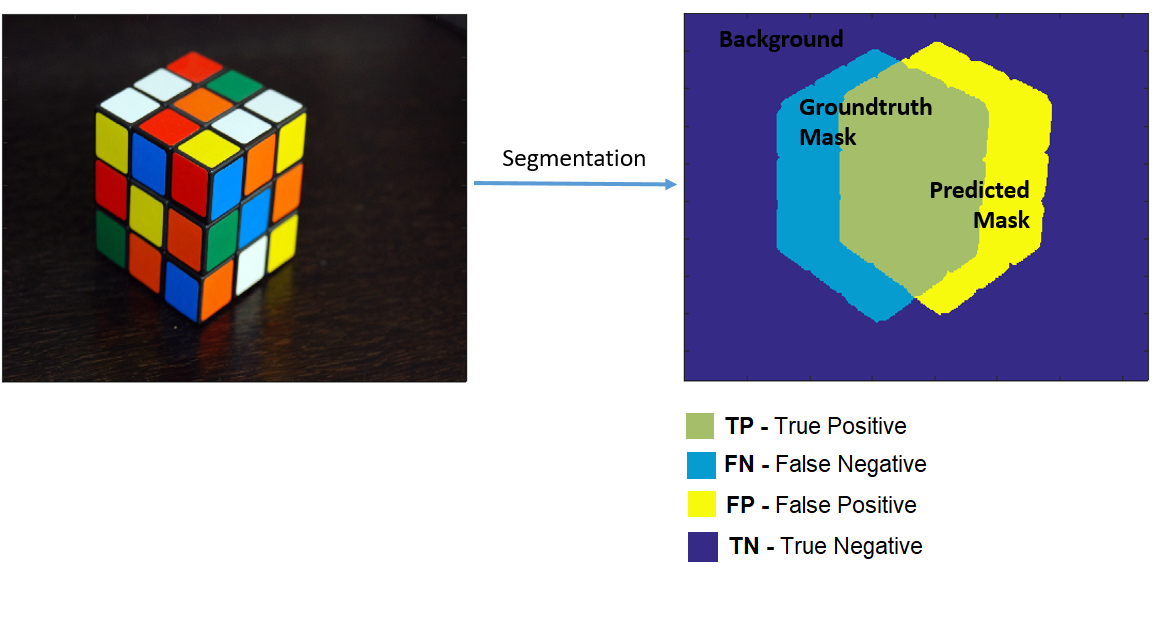}
    \caption{Visualization of the true positives, false positives and false negatives, given a predicted mask and its groundthruth mask.}
    \label{fig:eval_metrics}
\end{figure}

We use three main metrics, IoU (Intersection over Union), precision, and recall. Each metric is calculated for each image and averaged across all the images in the dataset. 
Given a predicted mask and its corresponding ground-truth mask, these metrics can be calculated as follows:

\begin{align}
    IoU &= \frac{TP}{TP + FP + FN} \\
    Precision &= \frac{TP}{TP + FP} \\
    Recall &= \frac{TP}{TP + FN},
\end{align}{}
where $TP$, $FP$, and $FN$ are pixels of the masks corresponding to the true positives, false positives, and false negatives, respectively, as shown in Figure~\ref{fig:eval_metrics}.

The evaluation metric used in the original work of Mask R-CNN \cite{R-cnn2017} is the mean Average Precision (mAP). 
This metric is widely used for tasks where we have multiple positive instances on the same image. These instances can belong to the same or to different classes, but the goal is to detect as many positive instances as possible while maintaining a reasonable accuracy in the masks and bounding box predictions. In the context of our work, this metric does not translate our goal and our specific problem since we typically have only 1 positive instance in each image and we want to have the highest accuracy possible when predicting its mask and bounding box. 
\section{Experiments and results}
\label{sec:results}

In this section, a series of experiments are conducted to produce both numerical (quantitative) and visual (qualitative) results, in order to test our hypotheses and compare different approaches. Our experiments aim to explore, in an empirical way, the effects of different training strategies and domain randomization parameters.
In Section~\ref{sec:dr_results}, we start by evaluate the effects of each randomization aspect presented in the simulated dataset generation (see Section~\ref{sec:Data_gen}).
Then, in Section~\ref{sec:lr_results} we will analyze what is the best solution for transferring the learned parameters on a generic dataset to our humanoid robotic hand segmentation task. In particular, which layers should be fine-tuned or frozen during training. 
Lastly, we will show the segmentation results on the simulated datasets and on the real-world datasets, showing quantitative and qualitative results.

\subsection{Domain Randomization results}
\label{sec:dr_results}

In this section, we show the results of training the model with different datasets generated in Unity. We train the \textit{model} on six different datasets, to measure the impact that some of the randomized aspects of the scene have on the accuracy of the \textit{model} when validated with real images. Then, we can chose the best model to use on test time. Moreover, the simulated validation datasets were used to chose when to stop the training process (early stop) and to be a base-line to the real validation set (\ie, how much performance are we losing by using domain randomization).

\begin{figure*}
    \centering
    \includegraphics[width=0.89\linewidth]{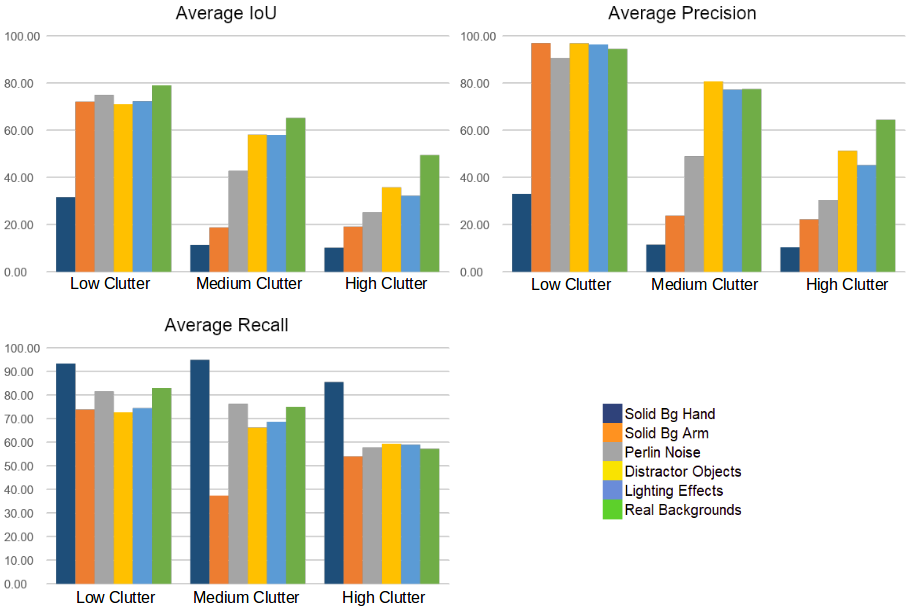}
    \caption{Results when validating models in different real validation datasets (namely, Low Clutter, Medium Clutter, and High Clutter), trained with different simulated datasets (namely, SolidBg Hand, SolidBg Arm, Perlin Noise, Distractor Objects, Lighting Effects, and Real Backgrounds). (Best seen in color)}
    \label{fig:dr_results}
\end{figure*}

In Figure \ref{fig:dr_results}, we present the results when validating the \textit{models} on each real validation dataset (Low, Medium and High Clutter), training the network using one of the six datasets generated in Unity (SolidBg Hand, SolidBg Arm, Perlin Noise, Distractor Objects, Lighting Effects, and Real Backgrounds, see Section~\ref{sec:simdatasets} for a description of the datasets). 

The hand and arm of the robot are very similar in color and texture when the back of the hand is turned to the camera. By training Mask R-CNN with a 3D model of the hand only (\textit{Solid Bg Hand} dataset - dark blue in Figure~\ref{fig:dr_results}) we can achieve a very high recall ($0.94, 0.95$ and $0.86$, on Low, Medium and High clutter datasets, respectively), but the \textit{precision} is very low ($0.33$, $0.12$ and $0.10$, respectively) due to parts of the arm visible in the image being detected as part of the hand. 

To avoid these false positives, the 3D model of the arm must be attached to the hand, in simulation, leading to a substantial improvement in \textit{precision}, with the dataset \textit{SolidBg arm} (orange in Figure \ref{fig:dr_results}), on all real validation datasets. The \textit{precision} using the \textit{SoliBg arm} dataset is $0.97$, $0.24$ and $0.22$, on the three validation datasets. However, this improvement comes with a small cost in \textit{recall}, making it harder for the model to detect the entirety of the hand. In general, adding the arm to the model is good since the average \textit{IoU}, which increased in all real validation datasets. From $0.32$ to $0.72$ on the Low Clutter dataset, from $0.12$ to $0.19$ on the Medium and from $0.10$ to $0.19$ on the High Clutter dataset.

The Perlin noise effect (using Perlin Noise dataset on training) gives the \textit{model} a capability of separating cluttered backgrounds from the hand. In Figure~\ref{fig:dr_results} and in grey color, we can see this effect improves the average \textit{IoU} in all real validation datasets, especially in the most cluttered ones. The \textit{IoU} on this new training dataset is $0.75$, $0.43$ and $0.25$, on the three validation datasets, respectively. This can be explained by the addition of randomness on the background which leads the model to reduce the detection of 'hands' on the background, \ie, False Positives (FP). Indeed, the precision increased by $0.25$ on the Medium Clutter and $0.09$ on the High Clutter datasets, only decreasing on the Low Cutter dataset by $0.06$.

This capability can also be extended by two important features. First, the addition of Distractor Objects using the Distractor Objects dataset (yellow in Figure.~\ref{fig:dr_results}) provides the training set with images where the hand is occluded by other objects. This also affects mostly the medium and high cluttered datasets which have images with different objects (in size, shape, and color) around and in front of the hand. This training set produces and average \textit{IoU} on the validation datasets of $0.71$, $0.58$ and $0.36$.

Secondly, by adding images with real backgrounds in the train set (Real Background dataset, in green in Figure~\ref{fig:dr_results}), we achieve the greatest performance in \textit{IoU} for the High cluttered dataset, \ie, $0.55$. The \textit{IoU} on the Low and Medium Clutter also increase compared to the previous training sets. This can be explained with the realistic negative examples on the background which reduce the False Negatives thus increasing the \textit{precision} of the \textit{model}.

However, when changing the lighting effects (using Lighting Effects dataset, in light blue in Figure~\ref{fig:dr_results}), we do not have significant differences in performance. We did not, however, explored many strategies in this regard. Future works could further explore this feature by introducing different types of lights with, for example, spotlight effects and place a different number of lights in different areas of the scene. 

\subsection{Transfer learning results}
\label{sec:lr_results}
In this Section~\ref{sec:lr_results} we will analyze what is the best solution for transferring the weights learned in a generic task (COCO) to our humanoid robotic hand segmentation task.

In order to conduct this experiment, we divide the network into 3 stages,
as explained earlier in Section~\ref{sec:mask_r-cnn_architecture}. The first stage is the Backbone, the second is the Proposal Stage, and finally, the Classification Stage (see Section~\ref{sec:mask_r-cnn_architecture} for more details on the network architecture and division)

For training and validation, we use the dataset with real backgrounds generated in Unity (see Section~\ref{sec:simdatasets} for a description of the datasets). 
For validation on real data, we merge our 3 real validation datasets into a single set and apply the evaluation metrics on it.

We train 4 \textit{models}, each initialized in a different way. The goal is to evaluate different combinations of initializations, by transferring \textit{weights}, pre-trained on COCO \cite{Lin2014}, to the given stage(s) and training the remaining stage(s) from scratch. The results of inferring the real and simulated validation datasets on these 4 \textit{models} are shown in Table \ref{tab:transfer_results}. 

\begin{table}
\caption{Results of inferring 4 \textit{models}, each trained from different initialization points, on the real and simulated validation datasets. The simulation datasets (training and validation) were the ones containing real backgrounds along with the other randomizations proposed. The Transferred weights were fine-tuned during the training process and the remaining were learn from scratch using a suitable initialization. Note that on the Classification stage the output layer is always learned from scratch since the number of classes changed.}
\label{tab:transfer_results}
\centering
\resizebox{\linewidth}{!}{%
\begin{tabular}{c c ccc c ccc}
\hline
& & \multicolumn{7}{c}{\textbf{Validation Sets}} \\
\cline{3-9} 
& & \multicolumn{3}{c}{\textbf{Simulation}} &                                         & \multicolumn{3}{c}{\textbf{Real}}                                                         \\ \cline{3-9} 
\multirow{-2}{*}{\textbf{Transferred weights}}     &  & \textit{Rec} & \textit{Pre} & \textit{IoU} & &\textit{Rec} & \textit{Pre} & \textit{IoU} \\ \hline
Backbone.                                                                  &  & 85,8                                 & 88,4                                 & 77,5
&
& 64,3                                 & 71,8                                 & 54,4                                 \\
\hline
\begin{tabular}[c]{@{}c@{}}Backbone and \\ proposal stage.\end{tabular}    &  & \textbf{90,0}                                 & 93,5                                 & \textbf{84,8}                         
&
& 62,5                                 & \textbf{86,9}                                 & 58,2                                 \\
\hline
\begin{tabular}[c]{@{}c@{}}Backbone and\\ classification stage.\end{tabular} & & 89,3                                 & \textbf{93,8}                                 & 84,5                                & & 55,2                                 & 68,1                                 & 50,8                                 \\
\hline
All stages.                                                                 & & 87,2                                 & 92,8                                 & 82,0                                & & \textbf{72,2}                                 & 79,5                                 & \textbf{63,4}                                 \\ \hline
\end{tabular}%
}
\end{table}

\begin{figure*}
    \centering
    \includegraphics[width=0.7\linewidth]{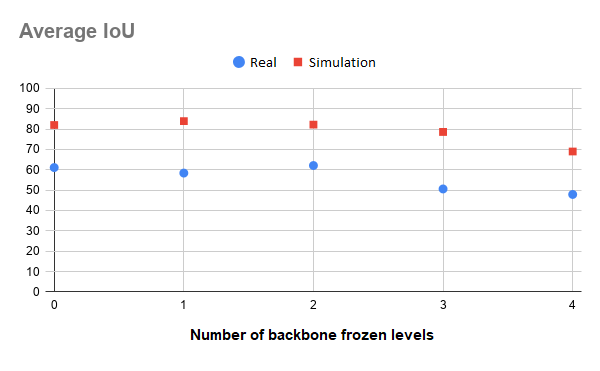}
    \caption{Results (average IoU) of inferring the real and simulated validation sets on 5 different \textit{models}, each trained with a number of backbone levels frozen.}
    \label{fig:tl_results}
\end{figure*}

Furthermore, we divide the backbone into 4 blocks of layers called levels, where the last layer of each block is the feed-forward connection to a different level of the FPN \cite{Lin2016} which, after a series of convolutional layers, results in 4 different feature map levels, to be used for region proposals and classification.
We train 5 different \textit{models}, each with a different number of backbone levels frozen during training. Given the sequential property of the backbone network, the number of levels frozen is counted from the input. In other words, when two levels are frozen, we mean that the weights of the first two levels of the network are not changed.
After this process, we apply the evaluation metrics (just the IoU, because now we are only interested in measuring the accuracy) to each \textit{model}, when inferring the real and simulated validation sets. The results of this experiment can be seen in Figure \ref{fig:tl_results}.
During this study, the remaining levels of the backbone and the other stages of the network (Proposal Stage and Classification) are fined-tuned transferring the weights learned in a generic task (COCO).

\subsection{Segmentation results}
\label{sec:inst_seg_results}

We train a \textit{model} with the configurations that yielded the best results, \textit{i.e.} we use the Real Background for training dataset - green dataset on Figure~\ref{fig:dr_results} .
We initialize all the stages of Mask-RCNN, except the output layers, with a model pre-trained on COCO and we fine-tune the network using a learning rate of $0.00025$. The learning rate was chosen by performing a line search within the range of values $[10^{-5}, 10^{-3}]$ and choosing the value that brings the best average accuracy, in the Real-World validation datasets. We only performed this search for the learning rate, due to the high computational cost of performing a grid search. Moreover, the learning rate is, usually, the \textit{hyperparameter} that brings the most variance to the results. We created a visualization of some results achieved by this \textit{model}, on real and synthetic validation images, by overlapping the predicted masks, bounding boxes and class scores on these images. 

\begin{figure*}
    \centering
    \subfloat[]{
    \includegraphics[width=0.495\linewidth]{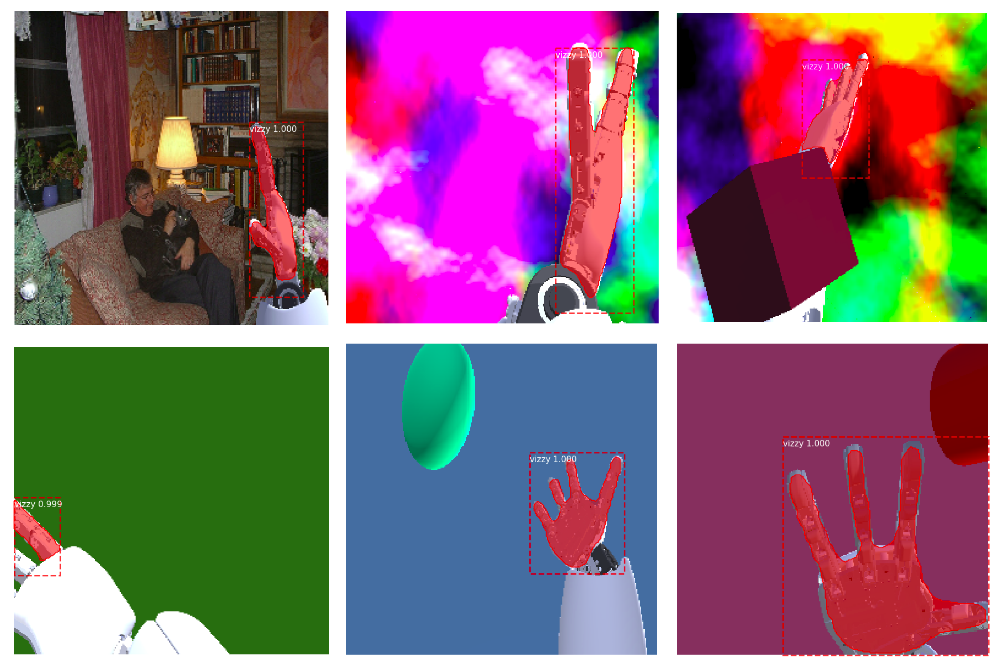}
    \label{fig:val_results}
    } 
    \subfloat[]{\includegraphics[width=0.495\linewidth]{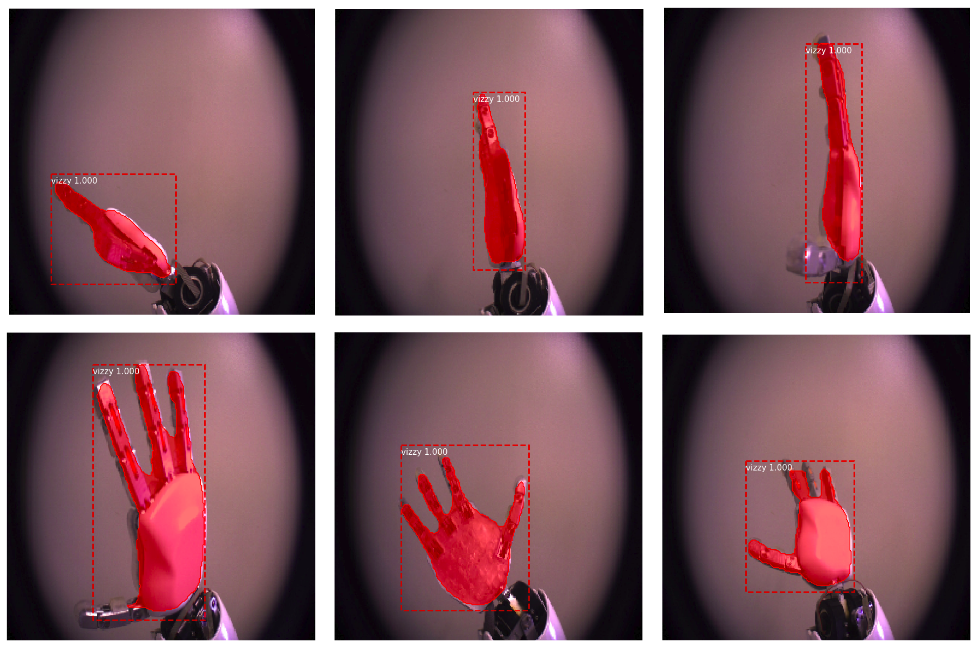}
    \label{fig:low_results}
    } \\
    \subfloat[]{
    \includegraphics[width=0.495\linewidth]{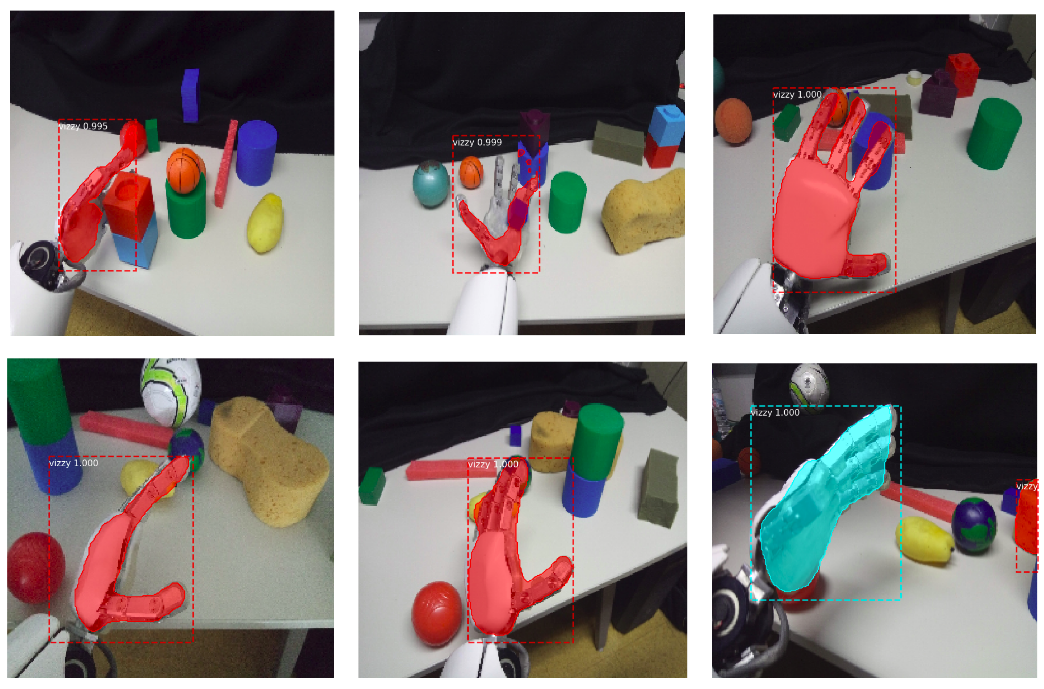}
    \label{fig:medium_results}
    }
    \subfloat[]{    \includegraphics[width=0.488\linewidth]{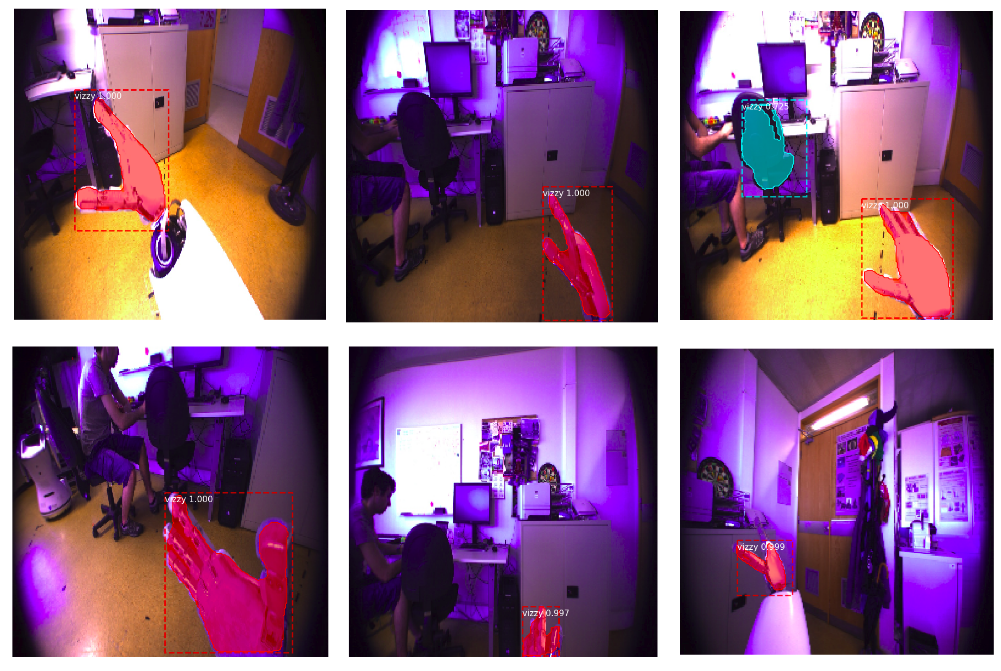}
    \label{fig:high_results}
    }
    \caption{Visualization of the predicted masks, bounding boxes and class scores on the validation datasets. (a) the simulated validation set, (b) the low clutter real validation set, (c) medium clutter real validation set and (d) the high clutter real validation set. (Best seen in color)}
\end{figure*}

We show visual results on the simulated validation (see Figure~\ref{fig:val_results}), Low Clutter (see Figure~\ref{fig:low_results}), Medium Clutter (see Figure~\ref{fig:medium_results}) and High Clutter (see Figure~\ref{fig:high_results}) datasets. The overall visual results are good, where most of the examples the hand are well detected and segmented, only missing some of the pixels on the boundaries. However, the model is far from perfect. Looking at the bad results, where the model fails the most, we can try to understand the reasons and sketch some future work. In Figure~\ref{fig:val_results}, bottom left, the model's prediction is poor due to the hand being in an unusual pose (\ie, a pose that was not sufficiently represented in the training set). Moreover, in Figure~\ref{fig:medium_results}, top middle, the model also shows poor accuracy when dealing with a partial occlusion of the hand. Another example of poor predictions is in Figure~\ref{fig:high_results}, top right. Although the \textit{recall} value of the predicted mask is very high (\ie, most of the hand's pixels were correctly predicted) the model also predicts a large portion of the background. All these results may be improved by introducing more balance in the training data, \ie, representing hand poses and background configurations more uniformly. Also, since the image view is egocentric, the hand of the robot cannot physically be positioned further than a certain distance from the camera. Therefore, we can also limit the position of the hand in the training set.

\begin{figure*}
    \centering
    \includegraphics[width=0.65\linewidth]{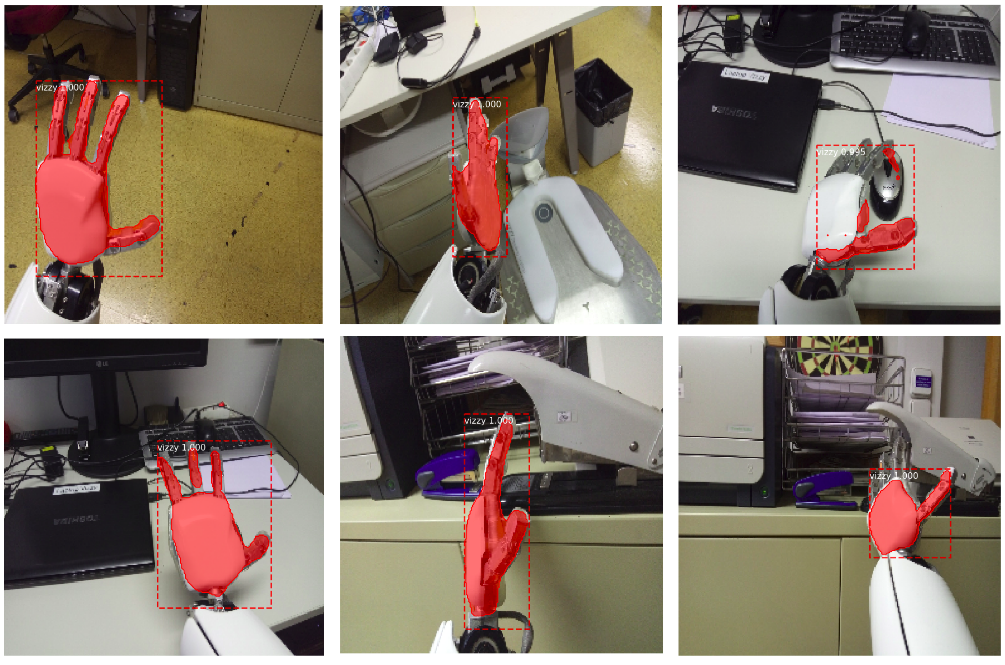}
    \caption{Visualization of the predicted masks, bounding boxes and class scores, for 6 images of the real test dataset. (Best seen in color)}
    \label{fig:test_results}
\end{figure*}

In order to test our final model's capacity to generalize to data that was neither used in training nor validation, we have used the real  test set dataset. As explain before, it was collected with the real robot in a scenario with the same level of difficulty of the High Clutter dataset. We inferred this dataset on our final model, yielding an average IoU of $0.563$. When comparing these results to the High Clutter dataset, where this model yield an average IoU of $0.554$, we can conclude that our model has the capacity of generalizing to unseen data.
The visual results can be seen in Figure~\ref{fig:test_results}. On the bottom right image it is possible to observe that the bounding box prediction does not contain the robot fingers. This is one of the problems that can arise from the proposed approach, \ie, first detecting the hand and then only run the segmentation process inside the detected bounding box. 

The results achieved by Lanillos \textit{et al.} \cite{lanillos2017tcds} are not directly comparable to ours since they explore robot motor commands to segment the robot body. However, it gives us an insight into the quality of our work which uses only vision perception, \ie, without any motor information. Comparing our results with the ones on Table~IV on \cite{lanillos2017tcds}, we can see that for a moving arm, the authors report an IoU of $0.615$. If the robot arm was static, they report a segmentation result of $0.513$.
We can state that our results, even without any motor information, are better in IoU accuracy.

\section{Conclusions and Future Work}
\label{sec:concl}

We aimed at giving a real robot the ability to perceive all the points of an image belonging to its own hand. This is done by fine-tuning a pre-trained CNN to extract binary masks of the robot's hand, from RGB images. To do this, we had to face two main challenges: (i) collect and annotate enough images, to train a deep CNN, and (ii) overcome the differences in the domains and tasks between the pre-trained model and our target model.

First, we developed a process for generating training and validation images, using Unity game engine and domain randomization concepts. Second, we applied transfer learning methods to the complex architecture of Mask R-CNN, in order to find a good strategy that fine-tunes the model to the new task.

With our approach, we were able to create a simple process for fine-tuning the \textit{weights} of a complex model using only, approximately, 1000 images for training and validation. 
The scene generation process is straightforward where each image contains a robotic hand, a background plane and, in some cases, distractor objects. Indeed, there is no need for data augmentation or other pre-processing methods. 
When evaluating our models in real environments we can retain most of the performance we get from evaluating in simulated environments. However, high cluttered environments can still significantly affect performance. 

Overall, we gave a detailed insight on different learning strategies and data generation methods, which can be used to train a state-of-the-art network (Mask R-CNN) and achieve the objective of this work, with a good performance. 
Moreover, we exploit vision sensing only to perform hand segmentation. Our proposed solution achieves an IoU accuracy better than the state-of-the-art, which explores motor commands as well.
However, it is a very time-consuming task to collect and annotate real images. For this reason, we were not able to create a real dataset large enough to make strong conclusions and evaluate the performance in many types of environments and with high variability in all environments' components.

Concerning the learning process, although the implementation of Mask R-CNN used allows for a large number of \textit{hyperparameters} to be tuned, we only tune the learning rate. If any future work has the possibility of training the network with several GPUs, at the same time, we recommend that a more detailed \textit{hyperparameter} search is made. Finally, another important addition to this work would be to include the information of the robot's kinematics, in the region proposal stage, in order to relieve the burden of generating proposals, from the network. With respect to the data generation process, future work can explore more randomization parameters and more suitable configurations for each parameter. Moreover, another possible direction to eliminate some background noise and to increase accuracy is to use depth information in order to not allow predictions detected further than a given distance from the camera view since the position of the robot hand is limited by its kinematics.


\bibliography{references}

\begin{thebibliography}{10}
\expandafter\ifx\csname url\endcsname\relax
  \def\url#1{\texttt{#1}}\fi
\expandafter\ifx\csname urlprefix\endcsname\relax\def\urlprefix{URL }\fi
\expandafter\ifx\csname href\endcsname\relax
  \def\href#1#2{#2} \def\path#1{#1}\fi

\bibitem{saponaro2017icdl}
G.~Saponaro, P.~Vicente, A.~Dehban, L.~Jamone, A.~Bernardino, J.~Santos-Victor,
  {Learning at the ends: From hand to tool affordances in humanoid robots}, in:
  Joint IEEE International Conference on Development and Learning and
  Epigenetic Robotics (ICDL-EpiRob), 2017, pp. 331--337.
\newblock \href {http://dx.doi.org/10.1109/DEVLRN.2017.8329826}
  {\path{doi:10.1109/DEVLRN.2017.8329826}}.

\bibitem{vicente2016jint}
P.~Vicente, L.~Jamone, A.~Bernardino, Robotic hand pose estimation based on
  stereo vision and gpu-enabled internal graphical simulation, Journal of
  Intelligent {\&} Robotic Systems 83~(3) (2016) 339--358.
\newblock \href {http://dx.doi.org/10.1007/s10846-016-0376-6}
  {\path{doi:10.1007/s10846-016-0376-6}}.

\bibitem{laflaquiere2019icdl}
A.~Laflaquière, V.~V. Hafner, {Self-supervised Body Image Acquisition Using a
  Deep Neural Network for Sensorimotor Prediction}, in: Joint IEEE
  International Conference on Development and Learning and Epigenetic Robotics
  (ICDL-EpiRob), 2019, pp. 117--122.

\bibitem{lanillos2017tcds}
P.~{Lanillos}, E.~{Dean-Leon}, G.~{Cheng}, {Yielding Self-Perception in Robots
  Through Sensorimotor Contingencies}, IEEE Transactions on Cognitive and
  Developmental Systems 9~(2) (2017) 100--112.
\newblock \href {http://dx.doi.org/10.1109/TCDS.2016.2627820}
  {\path{doi:10.1109/TCDS.2016.2627820}}.

\bibitem{vicente2016Frontiers}
P.~Vicente, L.~Jamone, A.~Bernardino, Online body schema adaptation based on
  internal mental simulation and multisensory feedback, Frontiers in robotics
  and AI 3 (2016) 7.

\bibitem{zenha2018icdl}
R.~Zenha, P.~Vicente, L.~Jamone, A.~Bernardino, {Incremental adaptation of a
  robot body schema based on touch events}, in: Joint IEEE International
  Conference on Development and Learning and Epigenetic Robotics (ICDL-EpiRob),
  IEEE, 2018, pp. 119--124.

\bibitem{stepanova2019ral}
K.~{Stepanova}, T.~{Pajdla}, M.~{Hoffmann}, {Robot Self-Calibration Using
  Multiple Kinematic Chains—A Simulation Study on the iCub Humanoid Robot},
  IEEE Robotics and Automation Letters 4~(2) (2019) 1900--1907.
\newblock \href {http://dx.doi.org/10.1109/LRA.2019.2898320}
  {\path{doi:10.1109/LRA.2019.2898320}}.

\bibitem{vicente2017icra}
P.~Vicente, L.~Jamone, A.~Bernardino, {Towards markerless visual servoing of
  grasping tasks for humanoid robots}, in: IEEE International Conference on
  Robotics and Automation, 2017, pp. 3811--3816.
\newblock \href {http://dx.doi.org/10.1109/ICRA.2017.7989441}
  {\path{doi:10.1109/ICRA.2017.7989441}}.

\bibitem{fantacci2017iros}
C.~{Fantacci}, U.~{Pattacini}, V.~{Tikhanoff}, L.~{Natale}, {Visual
  end-effector tracking using a 3D model-aided particle filter for humanoid
  robot platforms}, in: 2017 IEEE/RSJ International Conference on Intelligent
  Robots and Systems (IROS), 2017, pp. 1411--1418.
\newblock \href {http://dx.doi.org/10.1109/IROS.2017.8205942}
  {\path{doi:10.1109/IROS.2017.8205942}}.

\bibitem{nascimento2020icarsc}
M.~Nascimento, P.~Vicente, A.~Bernardino, {2D Visual Servoing meets
  Rapidly-exploring Random Trees for collision avoidance}, in: IEEE
  International Conference on Autonomous Robot Systems and Competitions
  (ICARSC), 2020.

\bibitem{Kakumanu2007}
P.~Kakumanu, S.~Makrogiannis, N.~Bourbakis, A survey of skin-color modeling and
  detection methods, Pattern Recognition 40 (2007) 1106--1122.
\newblock \href {http://dx.doi.org/10.1016/j.patcog.2006.06.010}
  {\path{doi:10.1016/j.patcog.2006.06.010}}.

\bibitem{Jones2002}
M.~J. Jones, J.~M. Rehg, {Statistical Color Models with Application to Skin
  Detection}, International Journal of Computer Vision 46~(1) (2002) 81--96.
\newblock \href {http://dx.doi.org/10.1023/A:1013200319198}
  {\path{doi:10.1023/A:1013200319198}}.

\bibitem{Khan2018}
A.~U. {Khan}, A.~{Borji}, Analysis of hand segmentation in the wild, in:
  IEEE/CVF Conference on Computer Vision and Pattern Recognition, 2018, pp.
  4710--4719.
\newblock \href {http://dx.doi.org/10.1109/CVPR.2018.00495}
  {\path{doi:10.1109/CVPR.2018.00495}}.

\bibitem{Leitner2013}
J.~Leitner, S.~Harding, M.~Frank, A.~Förster, J.~Schmidhuber, Humanoid learns
  to detect its own hands, IEEE Congress on Evolutionary Computation, CEC
  2013\href {http://dx.doi.org/10.1109/CEC.2013.6557729}
  {\path{doi:10.1109/CEC.2013.6557729}}.

\bibitem{R-cnn2017}
K.~He, G.~Gkioxari, P.~Doll{\'a}r, R.~B. Girshick, {Mask R-CNN}, IEEE
  International Conference on Computer Vision (ICCV) (2017) 2980--2988.

\bibitem{Lin2014}
T.-Y. Lin, M.~Maire, S.~Belongie, J.~Hays, P.~Perona, D.~Ramanan, P.~Dollár,
  C.~L. Zitnick, Microsoft coco: Common objects in context, Lecture Notes in
  Computer Science (2014) 740–755\href
  {http://dx.doi.org/10.1007/978-3-319-10602-1_48}
  {\path{doi:10.1007/978-3-319-10602-1_48}}.

\bibitem{Russakovsky2014}
O.~Russakovsky, J.~Deng, H.~Su, J.~Krause, S.~Satheesh, S.~Ma, Z.~Huang,
  A.~Karpathy, A.~Khosla, M.~Bernstein, A.~Berg, L.~Fei-Fei, Imagenet large
  scale visual recognition challenge, International Journal of Computer Vision
  115~(3) (2015) 211--252.
\newblock \href {http://dx.doi.org/10.1007/s11263-015-0816-y}
  {\path{doi:10.1007/s11263-015-0816-y}}.

\bibitem{Tobin2017}
J.~Tobin, R.~H. Fong, A.~Ray, J.~Schneider, W.~Zaremba, P.~Abbeel, Domain
  randomization for transferring deep neural networks from simulation to the
  real world, IEEE/RSJ International Conference on Intelligent Robots and
  Systems (IROS) (2017) 23--30.

\bibitem{Long2014}
E.~Shelhamer, J.~Long, T.~Darrell, Fully convolutional networks for semantic
  segmentation, IEEE Trans. Pattern Anal. Mach. Intell. 39~(4) (2017) 640--651.
\newblock \href {http://dx.doi.org/10.1109/TPAMI.2016.2572683}
  {\path{doi:10.1109/TPAMI.2016.2572683}}.

\bibitem{Farabet2013}
C.~{Farabet}, C.~{Couprie}, L.~{Najman}, Y.~{LeCun}, Learning hierarchical
  features for scene labeling, IEEE Transactions on Pattern Analysis and
  Machine Intelligence 35~(8) (2013) 1915--1929.
\newblock \href {http://dx.doi.org/10.1109/TPAMI.2012.231}
  {\path{doi:10.1109/TPAMI.2012.231}}.

\bibitem{Hariharan2014}
B.~Hariharan, P.~Arbel{\'{a}}ez, R.~Girshick, J.~Malik, {Simultaneous Detection
  and Segmentation}, in: D.~Fleet, T.~Pajdla, B.~Schiele, T.~Tuytelaars (Eds.),
  Computer Vision -- ECCV 2014, Springer International Publishing, Cham, 2014,
  pp. 297--312.

\bibitem{Dumoulin2016}
V.~Dumoulin, F.~Visin, A guide to convolution arithmetic for deep learning,
  arXiv preprint arXiv:1603.07285.

\bibitem{LinMS2016}
G.~{Lin}, A.~{Milan}, C.~{Shen}, I.~{Reid}, Refinenet: Multi-path refinement
  networks for high-resolution semantic segmentation, in: IEEE Conference on
  Computer Vision and Pattern Recognition (CVPR), 2017, pp. 5168--5177.
\newblock \href {http://dx.doi.org/10.1109/CVPR.2017.549}
  {\path{doi:10.1109/CVPR.2017.549}}.

\bibitem{RonnebergerFB15}
O.~Ronneberger, P.~Fischer, T.~Brox, {U-Net: Convolutional Networks for
  Biomedical Image Segmentation}, in: N.~Navab, J.~Hornegger, W.~M. Wells,
  A.~F. Frangi (Eds.), Medical Image Computing and Computer-Assisted
  Intervention -- MICCAI 2015, Springer International Publishing, Cham, 2015,
  pp. 234--241.

\bibitem{He2015}
K.~He, X.~Zhang, S.~Ren, J.~Sun, Deep residual learning for image recognition,
  in: IEEE Conference on Computer Vision and Pattern Recognition (CVPR), 2016,
  pp. 770--778.
\newblock \href {http://dx.doi.org/10.1109/CVPR.2016.90}
  {\path{doi:10.1109/CVPR.2016.90}}.

\bibitem{Pinheiro2016}
P.~O. Pinheiro, T.-Y. Lin, R.~Collobert, P.~Doll{\'{a}}r, {Learning to Refine
  Object Segments}, in: B.~Leibe, J.~Matas, N.~Sebe, M.~Welling (Eds.),
  Computer Vision -- ECCV 2016, Springer International Publishing, Cham, 2016,
  pp. 75--91.

\bibitem{Pinheiro2015}
P.~O. Pinheiro, R.~Collobert, P.~Doll\'{a}r, Learning to segment object
  candidates, in: Proceedings of the 28th International Conference on Neural
  Information Processing Systems - Volume 2, NIPS'15, MIT Press, Cambridge, MA,
  USA, 2015, pp. 1990--1998.

\bibitem{Girshick2015}
R.~Girshick, Fast r-cnn, in: IEEE International Conference on Computer Vision
  (ICCV), 2015, pp. 1440--1448.
\newblock \href {http://dx.doi.org/10.1109/ICCV.2015.169}
  {\path{doi:10.1109/ICCV.2015.169}}.

\bibitem{Lin2016}
T.-Y. Lin, P.~Doll{\'a}r, R.~B. Girshick, K.~He, B.~Hariharan, S.~J. Belongie,
  Feature pyramid networks for object detection, IEEE Conference on Computer
  Vision and Pattern Recognition (CVPR) (2017) 936--944.

\bibitem{Ken2002}
K.~Perlin, Improving noise, Proceedings of the 29th annual conference on
  Computer graphics and interactive techniques - SIGGRAPH '02 21 (2002) 681.
\newblock \href {http://dx.doi.org/10.1145/566570.566636}
  {\path{doi:10.1145/566570.566636}}.

\bibitem{YosinskiCBL14}
J.~Yosinski, J.~Clune, Y.~Bengio, H.~Lipson, {How transferable are features in
  deep neural networks?}, in: Z.~Ghahramani, M.~Welling, C.~Cortes, N.~D.
  Lawrence, K.~Q. Weinberger (Eds.), Advances in Neural Information Processing
  Systems 27, Curran Associates, Inc., 2014, pp. 3320--3328.

\bibitem{Ren2015}
S.~Ren, K.~He, R.~Girshick, J.~Sun, Faster r-cnn: Towards real-time object
  detection with region proposal networks, IEEE Transactions on Pattern
  Analysis and Machine Intelligence 39.
\newblock \href {http://dx.doi.org/10.1109/TPAMI.2016.2577031}
  {\path{doi:10.1109/TPAMI.2016.2577031}}.

\bibitem{Plinio2015}
P.~Moreno, R.~Nunes, R.~Figueiredo, R.~Ferreira, A.~Bernardino,
  J.~Santos-Victor, R.~Beira, L.~Vargas, D.~Arag{\~{a}}o, M.~Arag{\~{a}}o,
  {Vizzy: A Humanoid on Wheels for Assistive Robotics}, in: L.~P. Reis, A.~P.
  Moreira, P.~U. Lima, L.~Montano, V.~Mu{\~{n}}oz-Martinez (Eds.), Robot 2015:
  Second Iberian Robotics Conference, Springer International Publishing, Cham,
  2016, pp. 17--28.

\end{thebibliography}

\end{document}